\documentclass[10pt,twocolumn,letterpaper]{article}

\usepackage{cvpr}
\usepackage{times}
\usepackage{epsfig}
\usepackage{graphicx}
\usepackage{amsmath}
\usepackage{amssymb}
\usepackage{tabularx}
\newcommand*{\affaddr}[1]{#1} 

\newcommand*{\email}[1]{\texttt{#1}}
\usepackage[pagebackref=true,breaklinks=true,letterpaper=true,colorlinks,bookmarks=false]{hyperref}

\cvprfinalcopy 

\def\cvprPaperID{1895} 

\ifcvprfinal\fi
\begin{document}

\title{PuzzleNet: Scene Text Detection by Segment Context Graph Learning}
\author{%
	Hao Liu  \quad Antai Guo \quad  Deqiang Jiang \quad Yiqing Hu \quad Bo Ren\\		
	\affaddr{Tencent Youtu Lab} \quad\\
	\email{\small \{ivanhliu, ankerguo, dqiangjiang, hooverhu, timren\}@tencent.com}
}
\maketitle

\begin{abstract}
  Recently, a series of decomposition-based scene text detection methods has achieved impressive progress by decomposing challenging text regions into pieces and linking them in a bottom-up manner. However, most of them merely focus on linking independent text pieces while the context information is underestimated. In the puzzle game, the solver often put pieces together in a logical way according to the contextual information of each piece, in order to arrive at the correct solution. Inspired by it, we propose a novel decomposition-based method, termed Puzzle Networks (PuzzleNet), to address the challenging scene text detection task in this work. PuzzleNet consists of  the Segment Proposal Network (SPN) that predicts the candidate text segments fitting arbitrary shape of text region, and the two-branch Multiple-Similarity Graph Convolutional Network (MSGCN) that models both appearance and geometry correlations between each segment to its contextual ones. By building segments as context graphs, MSGCN effectively employs segment context to predict combinations of segments.  Final detections of polygon shape are produced by merging segments according to the predicted combinations. Evaluations on three benchmark datasets, ICDAR15, MSRA-TD500 and SCUT-CTW1500, have demonstrated that our method can achieve better or comparable performance than current state-of-the-arts, which is beneficial from the exploitation of  segment context graph.

\end{abstract}

\section{Introduction}
Scene text detection aims to accurately localize text in natural images. To date, a few works~\cite{zhou2017east,hu2017wordsup,liao2017textboxes,deng2018pixellink,long2018textsnake,lyu2018multi,ma2018arbitrary,liao2018rotation,wang2019shape,wang2019arbitrary,xie2019scene,baek2019character,wang2019shape,zhang2019look, Feng_2019_ICCV, Xing_2019_ICCV} have been proposed to address this task. However, scene text detection is still a complicated task due to three exclusive properties of scene text: First, scene text may appear in arbitrary irregular rotated shape, \textit{e.g.} curved form and trapezoid, in the scene image. Second, the lengths of different text lines have significant variations. Third, scene text could be characters, words, or text lines, which may confuse detection algorithm when determining the boundaries. 

\begin{figure}[t]
	\begin{center}
		\begin{tabular}{ccc}
			
			\includegraphics[width=0.95\linewidth]{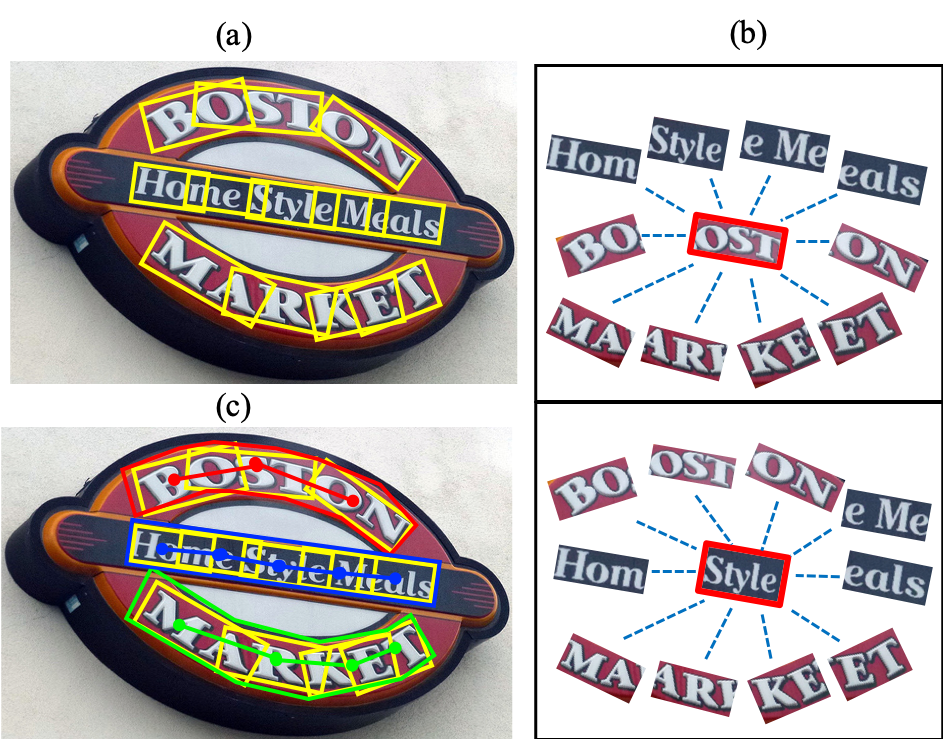}
			\vspace{-2mm}
			
		\end{tabular}
	\end{center}
	\caption{Illustration of the proposed framework, best viewed in color. Firstly, the oriented text segments (yellow  rectangles in (a) ) are detected,  some cases are skipped for better visualization.  Then, each segment plays as an anchor (highlighted by red borders in (b)) to be extracted correlations (blue dotted lines in (b)) between itself and its  contextual ones.  Finally,  the combinations between adjacent segment pairs are predicted (same-combination segments are connected by the same color solid lines in (c)). According to the predicted combinations, segments are merged into the final detection results (depicted in polygons of different colors in (c)).}
	\label{fig:motivation}
	\vspace{-1.5mm}
\end{figure}

Alternatively, one category of prevalent methods is decomposition-based methods~\cite{tian2016detecting,hu2017wordsup,shi2017detecting,baek2019character, Feng_2019_ICCV, Xing_2019_ICCV}, which handle challenging texts by decomposing them into pieces and linking those belonging to the same text region in a bottom-up manner. However, most of these previous methods simply focus on linking individual text pieces while the context information of each text piece is not given sufficient attention. In real situations, people are likely to judge whether two text pieces belong to the same text region according to their context information (\textit{e.g.,} character layout or font), rather than the limited information carried by two individual pieces themselves.  This is similar to puzzle game, in which the solver put pieces together according to the  patterns and  particular ordering of contextual pieces, in order to arrive at the correct solution.


 Inspired by above observations, we propose a novel decomposition-based scene text detection architecture, named Puzzle Networks (PuzzleNet). The overview of the proposed framework is illustrated in Fig.~\ref{fig:motivation}. We decompose arbitrary-shape text region into a number of text pieces represented by \textit{segments}. Here, a segment refers to an oriented rectangle (yellow rectangle in (a)) covering a part of the text region. Then each segment is regarded as an anchor (highlighted by red borders in (b)) to model the interior correlations (blue dotted  lines in (b)) between it and contextual segments. Then segments are merged together to form the final detection results (polygons of different colors in (c)) according to the predicted combinations (same-combination segments are connected by the same color solid lines in (c)). Here, segments from the same text region are deemed in a correct \textit{combination}. 
 
 The above text detection procedure involves two non-trivial problems: 1) predicting high-quality candidate text segments for the following context construction; 2) integrating correlations between each segment and contextual ones into combination prediction procedure.
 
   Most of previous works~\cite{shi2017detecting, hu2017wordsup, baek2019character, Feng_2019_ICCV, Xing_2019_ICCV} represent text pieces as orientated square boxes or character boxes. This type of representation has a main drawback that the covering region of each independent box is too limited to provide sufficient information for linking independent box pairs together. Moreover, we observe that the adjacent characters in a local region often have similar orientations, though different characters in curved shape text have different orientations. Take the curved-line text ``BOSTON'' in Fig.~\ref{fig:motivation}(a) for example, the region of ``BO'' has an approximately same orientation while the region of ``OST'' has another one. Therefore, in this work we propose to represent text piece by rectangular segment covering larger local text region with more effective information. Instead of greedily predicting  orientated square box at every position of text region, a more effective way is to predict the fewer number of segments covering text regions with similar orientations. To achieve this goal,  we propose the local orientation aware Segment Proposal Network (SPN) to detect segments.

The remaining question is how to make full use of segments and their context to predict their combinations. We propose to construct segments as context graphs, in which each node corresponds to a segment. Correlations between segments are represented by edges. In this way, each segment can be either an anchor or one of context to others simultaneously. Aiming at inferring combinations of each adjacent pairwise segments, we coin the innovative two-branch Graph Convolution Network (GCN) to perform reasoning on segment context graphs from both appearance and geometry perspectives.  Furthermore, we extend GCN to the Multiple-Similarity GCN (MSGCN) by introducing multiple-similarity mechanism to better capture the appearance and geometry correlations between segments.  In the end, corresponding segments belonging to the same combination are grouped into several clusters  and merged to form the polygon shape of detection results.


\begin{figure*}[htb]
	\centering
	\includegraphics[width=13cm]{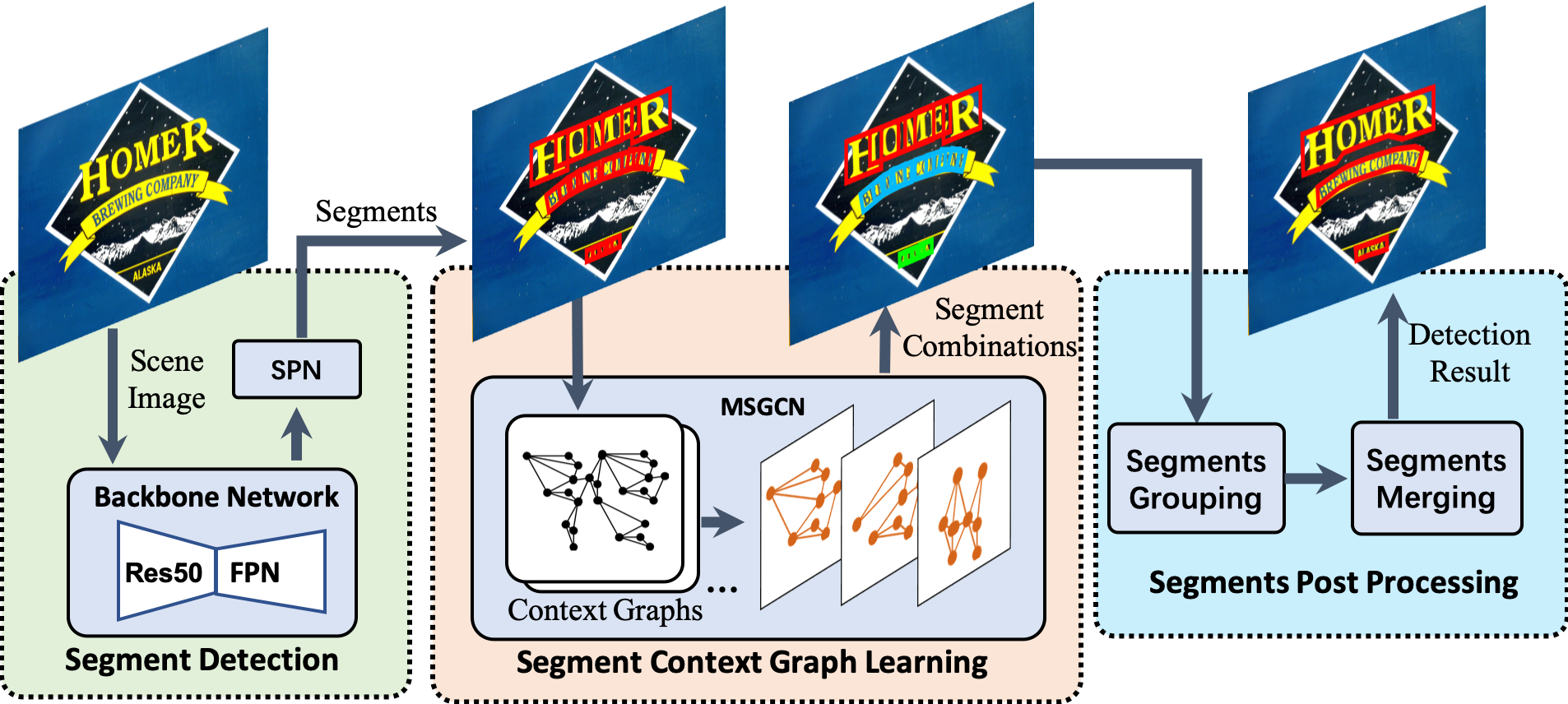}
	\caption{The overview of our proposed Puzzle Networks (PuzzleNet). Given a scene image, the network outputs text segments and segment combinations by segment detection and Multiple-Similarity Graph Convolutional Network (MSGCN). Finally,  detection result is generated by grouping and merging segments. Best viewed in color.} 
	
	\label{fig:arch}
\end{figure*}

Our contributions are in the following three folds: 1) A novel scene text detector named PuzzleNet is proposed, which is flexible for text instance with arbitrary shape, orientation and varying aspect ratio. 2) A local orientation aware SPN is designed to predict the segments with more effective representation of text regions. 3) To our best knowledge, we are the first to build segments as context graphs and learn their correlations considering context information for predicting their combinations. We evaluate our approach on three public benchmarks, ICDAR15, MSRA-TD500 and SCUT-CTW1500. Experimental results demonstrate that our method is able to achieve significantly better performance compared with state-of-the-arts. 
\section{Related Work}

With the development of deep learning, scene text detection has recently advanced substantially on both performance and robustness. The deep learning-based scene detection methods can be roughly categorized into three groups: regression-based, segmentation-based and decomposition-based methods. 

\textbf{Regression-based methods}~\cite{liao2017textboxes,zhou2017east,ma2018arbitrary,liao2018rotation} aims to design the robust generation method of bounding box to improve the performance. TextBoxes~\cite{liao2017textboxes} introduce long default boxes and convolutional filters to cope with the long aspect ratio text lines. In \cite{ma2018arbitrary}, the Rotation Region Proposal Networks (RRPN) are designed to generate inclined proposals with text orientation angle information. Besides, Liao \textit{et al.}\cite{liao2018rotation} propose to perform classification and regression on features of different characteristics with a two-branch model. However, these methods are still disturbed by the large aspect ratio variations and irregular shape of the text region. 


\textbf{Segmentaion-based methods}~\cite{yao2016scene,deng2018pixellink,long2018textsnake,wang2019shape,xie2019scene,zhang2019look} regard all the pixels within text bounding boxes as positive regions and directly draw text bounding boxes from predicted segmentation map.  
Pixellink~\cite{deng2018pixellink}  predicts instance-segmentation map and linking pixels within the same instance together. However, this method is liable to failure when two text lines lie close to each other. Recent work~\cite{lyu2018multi} introduce position-sensitive segmentation to solve the problem. But it can only handle the rectangular shape text region.  Textsnake~\cite{long2018textsnake},  PSENet~\cite{wang2019shape} , LOMO~\cite{zhang2019look} and SPCNET~\cite{xie2019scene} predict text center line map to separate different text instances, whose performance strongly affected by the robustness of segmentation results.

\textbf{Decomposition-based methods}~\cite{tian2016detecting,hu2017wordsup,tian2017wetext,shi2017detecting,baek2019character, Feng_2019_ICCV, Xing_2019_ICCV} first decompose text region into pieces or characters, and then group them into final detection results. Our method belongs to this category which can essentially well handle the text cases with arbitrary shape, orientation and aspect ratio. Among the recent methods, CTPN~\cite{tian2016detecting} generates dense and compact text components. In character level, WordSup~\cite{hu2017wordsup}, Wetext~\cite{tian2017wetext} and CRAFT~\cite{baek2019character} propose to localize individual characters in the image and group them  into a single instance. Although introducing weakly supervised learning, these methods still need strict character level annotations. 

Particularly, different from SegLink~\cite{shi2017detecting} using the square shape segment, our method focuses on detecting the arbitrary-shape text and predicting rectangular segments which carry more effective information. Comparatively, SegLink can only detect the text in line shape by predicting square boxes. Besides, all above mentioned decomposition-based methods only exploit the information carried by individual text piece pairs themselves when predicting the relationship or affinity between them, in which context information of text pieces, such as character layout structure, is not fully utilized. Contrastively, our proposed PuzzleNet can capture and aggregate the context information of segments by building segment context graphs to facilitate the prediction of their combinations.



\section{Proposed Puzzle Networks}\label{Pzznet}


The overview of the proposed Puzzle Networks (PuzzleNet) is shown in Fig.~\ref{fig:arch}. It consists of three components, \textit{i.e.}, segment detection, segment context graph learning and segment post processing. First, segments are predicted based on the proposed local orientation aware Segment Proposal Network (SPN). After building them as context graphs, we perform reasoning on graphs via the newly proposed Multiple-Similarity Graph Convolutional Network (MSGCN) to predict segment combinations. Finally, the segments are grouped into several clusters according to the predicted combinations, and then the segments in the same cluster are merged as the final detection result.

%

\subsection{Backbone Network}

Our method adopts ResNet~\cite{he2016deep}  of 50 layers with a Feature Pyramid Network (FPN)~\cite{lin2017feature} as the backbone network (illustrated in green part of Fig.~\ref{fig:arch}). For a single-scale input scene image, FPN utilizes a top-down architecture with lateral connections to build an feature pyramid from it. Features output by last residual blocks of  \textit{conv2}, \textit{conv3}, \textit{conv4} and \textit{conv5}  are denoted as \{\textit{$C_2$}, \textit{$C_3$}, \textit{$C_4$}, \textit{$C_5$}\}, which respectively have strides of \{4, 8, 16, 32\} pixels with respect to the input image. Correspondingly, a set of upsampled features output by FPN is called \{\textit{$P_2$}, \textit{$P_3$}, \textit{$P_4$}, \textit{$P_5$}\}. Note, slightly different from the vanilla ResNet50-FPN in \cite{lin2017feature}, we insert non-local blocks proposed in \cite{wang2018non}  after  \textit{$C_2$} and \textit{$C_4$}  to capture long range dependencies, which can model interactions between any two pixels, regardless of their positional distance. 
\subsection{Segment Detection}\label{sec:SD}
 As discussed, our method first detects a set of segments satisfying that local adjacent text regions with a similar orientation covered by a rectangular segment. To this end, we design the local orientation aware Segment Proposal Network (SPN) and generate the Ground Truth (GT) segment from GT polygon for training it. In our method, the segment is denoted as $\textit{S} = (x,y,w,h,\theta)$, 
where $(x,y)$ is the center point of segment while height $h$ and width $w$ correspond to its short side and long side respectively. $\theta$ represents the angle measured counter-clockwise from the positive x-axis. 
\begin{figure}[t]
	\begin{center}
		\begin{tabular}{ccc}
			
			\includegraphics[height=0.5\linewidth]{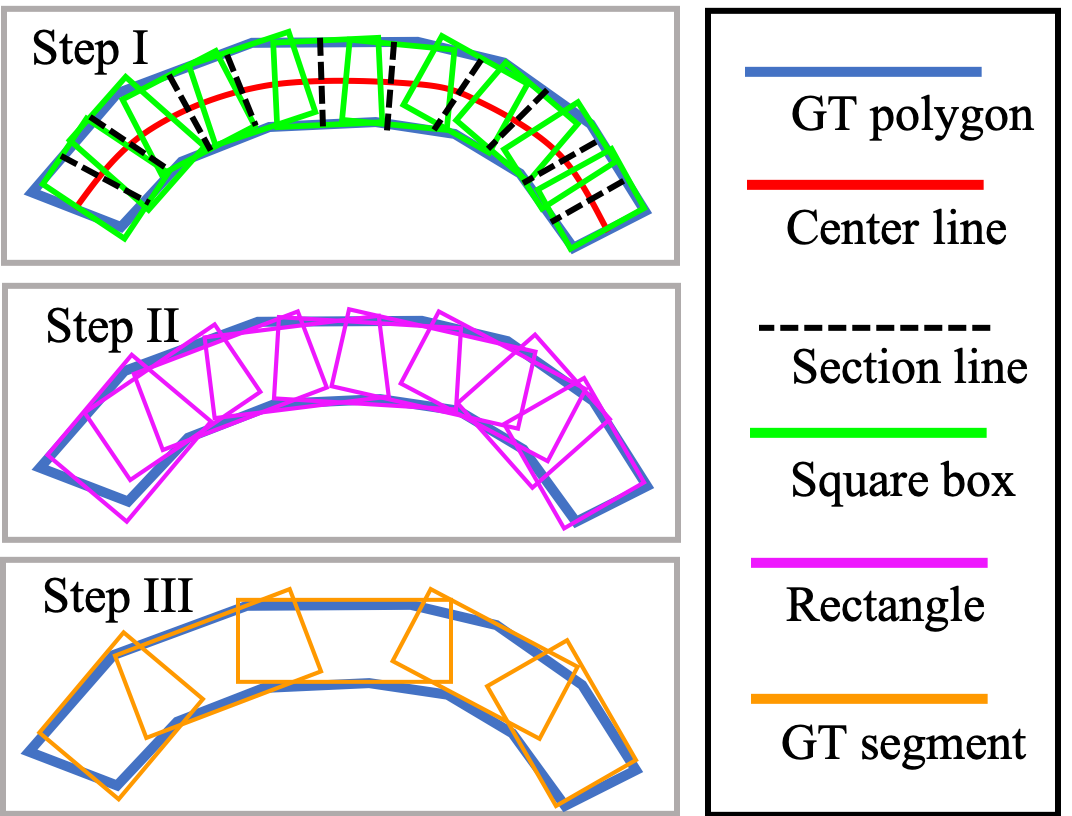}
			\vspace{-3mm}
			
		\end{tabular}
	\end{center}
	\caption{Illustration of ground truth (GT) segment generation. The procedure of generation includes three steps: I. Extracting text center line and generating square boxes covering GT polygon; II. Merging square boxes into rectangles; III. Merging rectangles with similar orientations into GT segments. Best viewed in color.} 
	\label{fig:lab}
	\vspace{-3mm}
\end{figure}

\textbf{Ground Truth Segment Generation}\quad We first elaborate on how to generate Ground Truth (GT) segment from GT polygon as supervision. More concretely, we decompose the GT polygon into several GT segments in three steps, which is illustrated in Fig.~\ref{fig:lab}. It is easy to directly decompose rectangular box into segments. However, decomposition on polygons of more than 4 sides is not easy with a general algebraic method. Therefore, we firstly adopt the method in \cite{long2018textsnake} to generate the text center line (red solid line in Fig.~\ref{fig:lab}) of polygon (depicted in dark blue in Fig.~\ref{fig:lab}) . Then the center line is evenly sectioned into $n$ line segments by $n-1$ section points. We set $n$ to 50 in this paper. The interval $\tau$ is determined by the following equation:
\begin{equation}
\small
\tau = \sigma\cdot\min(\lbrace\| l^{cross}_i \|\rbrace), i = 1, 2,...,n-1.
\end{equation} 
And $\|l^{cross}_i \|$ denotes the length of $i$-th cross section line (doted line in Fig.~\ref{fig:lab}) at $i$-th sample point. $\sigma$ is the scale coefficient which is set to 0.5. Then, each cross section line plays as the axis to generate a square box (green box in Fig.~\ref{fig:lab}) covering part of polygon. The side length of the square box equals to the length of corresponding cross section line.
In the second step, each two neighbouring square boxes are merged into a rectangle (magenta boxes in Fig.~\ref{fig:mg}) by calculating the enclosing rectangle of region covered by two square boxes. In the final step, those adjacent rectangles with angle difference smaller than $\pi/36$ are further merged into longer ones as GT segments (orange boxes in Fig.~\ref{fig:lab}) if the aspect ratios of them suffice to smaller than 3. By this way, the GT segments with local similar orientations are generated. 

\begin{figure*}[htb]

			\includegraphics[width=0.96\linewidth]{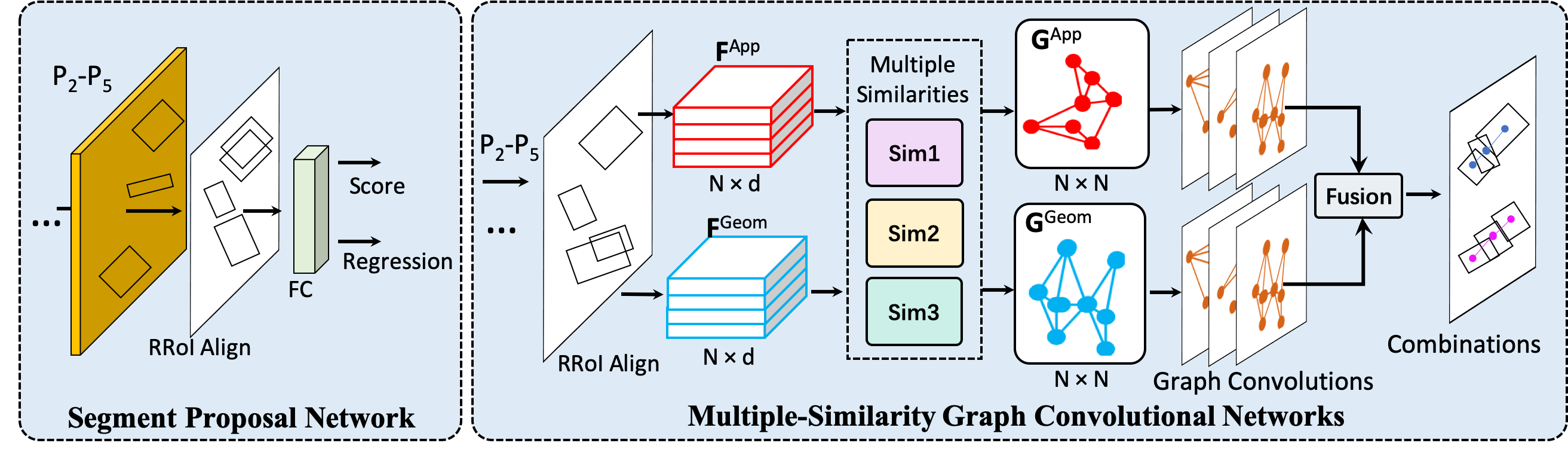}
			
			
	\caption{The detailed network of Segment Proposal network(SPN) and Multiple-Similarity Graph Convolutional Network (MSGCN). Best viewed in color.} 
	\label{fig:dt}
	\vspace{-1.5mm}
\end{figure*}

\textbf{Segment Proposal Network}\quad The detailed structure of Segment Proposal Network (SPN) is depicted in Fig.~\ref{fig:dt}, which is derived from Rotation Region Proposal Network (RRPN)~\cite{ma2018arbitrary}. To expediently detect segment, in SPN we use 4 scale anchors  of $\{32^2, 64^2 , 128^2, 256^2 \}$ with multiple orientations ${\{-\pi/6, 0, \pi/6, \pi/3, \pi/2, 2\pi/3\}}$ on \{\textit{$P_2$}, \textit{$P_3$}, \textit{$P_4$}, \textit{$P_5$}\} respectively. The aspect ratios are set to $\{1.5, 2, 2.5\}$. Further, we assign RRoIs of different sizes to different pyramid levels with different scales. The assign rule inherits from \cite{lin2017feature}. Besides, instead of using RRoI pooling~\cite{ma2018arbitrary}, we apply RRoI Align~\cite{huang2018improving} on different level RRoIs to avoid misaligned results. The features of $7 \times 7 \times d$ size output by RRoI Align are then sent to two Fully-Connected (FC) layers with 512 dimensions. The segment score and coordination regression are then performed via linear layers.



\subsection{Segment Context Graph Learning}\label{SCP}
In this section, we introduce the detailed structure of the proposed segment context graph as well as the Multiple-Similarity Graph Convolutional Network (MSGCN) to learn graph parameters. The details is illustrated in Fig.~\ref{fig:dt}.

\textbf{Multiple-Similarity Graph Convolutional Network}  Given a set of segments, the objective of MSGCN is to construct them as graphs to jointly take the segments and their context information into consideration,  and judge whether segment pairs belong to the same combination.

We observe that segments belonging to the same combination often have similar appearance (\textit{e.g.,} character font)  and their layouts have certain natural graph structures. Therefore, we construct segments as graphs from both \textit{appearance} and  \textit{geometry} perspectives. In the constructed graphs, each segment can be either an anchor or one of context of others.
In particular, we first extract their \textit{appearance feature} and \textit{geometry feature} as nodes to construct graphs. 

To extract appearance feature,  the RRoI Align is applied on corresponding pyramid level in the similar way with segment detection stage.  Three convolutional layers with $3 \times 3 \times 512 $ size of kernel are then applied. We denote the  segment number as $N$. After passing a FC layer with $d$ dimensions,  appearance features $\mathbf{F}^{\text{App}} = \left\lbrace \mathbf{f}_1, \mathbf{f}_2, ...,\mathbf{f}_N \right\rbrace \in \mathbb{R}^{N \times d}$ are obtained, as depicted in red cubes in Fig.~\ref{fig:dt}.  

Basing on the segment denotion in Sec.~\ref{sec:SD}, we  derive the geometry feature of each segment as 
$\left({\frac{x}{W}}, {\frac{y}{H}}, {\frac{w}{W}}, {\frac{h}{H}}, \theta \right)^\top$, where $W$ and $H$ are the width and height of the scene image. Then a $d$-dimension FC is applied on above vectors to obtain the geometry features $\mathbf{F}^{\text{Geom}} = \left\lbrace \mathbf{g}_1, \mathbf{g}_2, ...,\mathbf{g}_N \right\rbrace \in \mathbb{R}^{N \times d}$, which described in blue cubes in Fig.~\ref{fig:dt}.

Then, two kinds of nodes are connected by two types of edges separately: \textit{appearance similarity} and \textit{geometry similarity}. With similarity relations, we can model the interplay of appearance and geometry between each segment and its contexts. The appearance similarity $\mathbf{G}^{\text{App}}$ (red graph in Fig.~\ref{fig:dt}) and geometry similarity $\mathbf{G}^{\text{Geom}}$ (blue graph in Fig.~\ref{fig:dt}) can be computed as
\begin{align}
\small
&\mathbf{G}_{i,j}^{\text{App}} = K(\mathbf{f}_i,\mathbf{f}_j), i, j \in \{1,2,..., N\},\label{App}\\
&\mathbf{G}_{i,j}^{\text{Geom}} = K(\mathbf{g}_i,\mathbf{g}_j) , i, j \in \{1,2,..., N\},\label{Geom}
\end{align}
where $K$ is the similarity function. To better capture the correlations between segment pairs, we equip it with three types of similarities, which can be represented as
\begin{align}
\small
&K = \beta_1K_1 + \beta_2K_2 + \beta_3K_3 , \text{s.t. }   \beta_1 + \beta_2 +\beta_3 = 1,\\
&K_1(\textbf{y}_1,\textbf{y}_2) = \frac{\textbf{y}_1 \cdot \textbf{y}_2^\top} {( \left| \textbf{y}_1\right| \cdot \left| \textbf{y}_2\right|)},\\
&K_2(\textbf{y}_1,\textbf{y}_2) = \exp(-\frac{\left\|\textbf{y}_1 - \textbf{y}_2\right\|^2}{2\sigma^2}),\\
&K_3(\textbf{y}_1,\textbf{y}_2) = \exp(-\frac{\textit{JSD}(\textbf{y}_1,\textbf{y}_2)}{2\sigma^2}),\\
&\textbf{y}_1 = \phi_1(\textbf{x}_1) = \textbf{w}_1\textbf{x}_1,\quad \textbf{y}_2 = \phi_1(\textbf{x}_2) = \textbf{w}_2\textbf{x}_2.
\end{align} 
$K_1$, $K_2$ and $K_3$ are cosine similarity, Gaussian similarity and Jensen-Shannon similarity~\cite{bai2013graph} respectively. \textit{JSD} represents the Jensen-Shannon Divergence and $\sigma$ is the control parameter set to 5 in this work. $(\beta_1, \beta_2, \beta_3)$ are similarity weight parameters which can be learned through back propagation. $\phi_1$ and $\phi_2$ represents two different transformations of the input features $\textbf{x}_1$ and $\textbf{x}_2$.  The parameters $\textbf{w}_1$ and $\textbf{w}_2$ are both $d \times d$ dimensions weights which can be learned via back propagation. Note, $\mathbf{G}^{\text{App}}$ and $\mathbf{G}^{\text{Geom}}$ are produced by passing 
$\mathbf{F}^{\text{App}}$ and $\mathbf{F}^{\text{Geom}}$ through $K$ individually. By introducing multiple-similarity mechanism to  our graph reasoning model,  it enables more effective message passing between each node. It also provides significant boost to the performance of our scene text detection task which will be verified by analytic experiments in Sec.~\ref{sec:abs}. Then the $\mathbf{G}^{\text{App}}$ and $\mathbf{G}^{\text{Geom}}$  are adopted as the adjacency matrix representing the similarity graphs in our work. 

To perform reasoning on context graphs, the Graph Convolutional Network (GCN)~\cite{kipf2016semi} is applied. Formally, one basic layer of graph convolutions can be represented as $\mathbf{Y} = \mathbf{G}\mathbf{X}\mathbf{W}$,
where $\mathbf{G} \in \mathbb{R}^{N \times N}$ is the adjacency matrix we have introduced above ($\mathbf{G}^{\text{App}}$ or $\mathbf{G}^{\text{Geom}}$), $\mathbf{X} \in \mathbb{R}^{N \times d}$ is the input features, $\mathbf{W}$ represents the $d \times d$-dimension weight matrix while $\mathbf{Y} \in \mathbb{R}^{N \times d}$ denotes the output features. To make the information flow unimpeded, we equip every layer of GCN with a residual connection in our method. Then the graph convolutional layer can be extended as 
$\mathbf{Y} = \mathbf{G}\mathbf{X}\mathbf{W} + \mathbf{X}.$

Then, the Layer Normalization and ReLu layers are applied on the output of each layer. In this work, we apply two branches of 3-layer GCN  on $\mathbf{G}^{\text{App}}$ and $\mathbf{G}^{\text{Geom}}$ individually, as illustrated in the Fig.~\ref{fig:dt}. After that, the outputs of two branches are fused by concatenating them into the feature in $N \times 2d$ dimensions. To predict final combination scores of each pairwise nodes, we apply two $N$-dimension FC layers and a convolutional layer with $1 \times 1 \times 2$ kernel size on the fused feature to obtain the feature map in $N \times N \times 2$ dimensions. In the end, the position-wise softmax function is performed on it to get the final combination score map $\mathbf{C}$. The score $s_{i,j} \in \mathbf{C}$ indicates the probability of $i$-th segment and $j$-th segment belonging to the same combination.

\subsection{Segment Post Processing}\label{spp}
After segments and corresponding combination score maps obtained, two steps of post processing should be performed to output the final detection results, as shown in Fig.~\ref{fig:arch}. 

\textbf{Segment Grouping}  
The first step is grouping candidate segments into different clusters according to the combination score maps. Our model infers the combinations of the adjacent segment pairs. To construct pairs, we exploit \textit{k}-Nearest Neighbors with Radius (kNNR)~\cite{yu2010high} algorithm for each segment. For $m$-th segment $S_m$, its adjacent pair segments $S_n \in \mathbf{S}^{\text{adj}}$ should subject to the following two rules: (1) $S_n$ are in the set of \textit{k}-Nearest Neighbors, which are determined by the Euclidean distance of their centers $D_{m,n} = \sqrt{(x_m-x_n)^2+(y_m-y_n)^2}$; (2) $D_{m,n} < R$, where $R$ is the radius computed as $R = \alpha\sqrt{w_m^2+h_m^2}$. In this work, $k$ and $\alpha$ are set to 5 and 2.5 respectively. Then, for each segment $S_m$, its matchable segment $\hat{S}$ (linked by the same color lines in Fig.~\ref{fig:motivation}) should satisfy that $\hat{s} = max(\mathbf{C}^\text{adj})$ and $\hat{s}>0.5$, where $\hat{s}$ is the combination score of $\hat{S}$. Through this way, all  segments can be grouped into several clusters. As is shown in Fig.\ref{fig:arch}, three clusters of segments are represented in red, blue and green respectively.

%

\begin{figure}[t]
	\begin{center}
		\begin{tabular}{ccc}
			{\hspace{-3pt}}
			
			\includegraphics[height=0.34\linewidth]{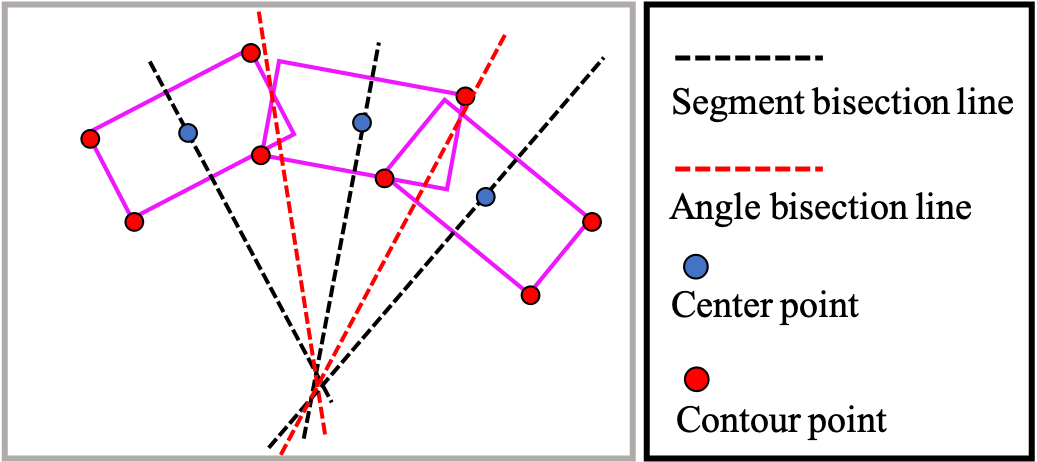}
			\vspace{-3mm}
			
		\end{tabular}
	\end{center}
	\caption{Illustration of segment merging procedure. Best viewed in color. } 
	\label{fig:mg}
	\vspace{-3mm}
\end{figure}

\textbf{Segment Merging} 
In each cluster,  we firstly generate segment bisection line (black dotted line in Fig.~\ref{fig:mg}) of each segment along short side direction, then the angle between two neighbouring center lines are evenly divided by a bisection line (red dotted line in Fig.~\ref{fig:mg}). After that, the vertices on the closest short sides of two segments are extracted, denoted as $\{{v^{(e)}_{f}}\} , e,f \in \{1,2\}$,  where $e$ stands for the $e$-th segment and $f$ is the $f$-th vertex. Correspondingly, we project the vertices on the angle bisection line and two projected points with max distance are kept. Their corresponding vertices are regarded as \textit{contour points}(red points in Fig.~\ref{fig:mg}). Note, if two segments have the same angle, the center line of any one of them is deemed as bisection line. Finally, the contour points are connected along the direction of segment long side. To make the boundary curve of final detection more smooth, a thin plate spline (TPS)~\cite{bookstein1989principal} method is applied.

\section{Training Strategy}

We train our proposed PuzzleNet in an end-to-end way. And the multi-task loss function is defined as
\begin{equation}
\small
L = L_{reg} + \lambda_{1} L_{score}+ \lambda_{2}L_{comb},
\end{equation} 
where $L_{reg}$ and $L_{score}$  are the loss functions of  SPN while $L_{comb}$ is the ``combination'' loss of combination prediction in MSGCN.  $\lambda_{1}$ and $\lambda_{2}$  are the weights constants to control the trade-off of three terms. In default, we set the $\lambda_{1}$ and  $\lambda_{2}$ to 1 and 5. 
We regress the predicted tuple of text label $\mathbf{v} = (v_{x}, v_{y}, v_{w}, v_{h}, v_{\theta} )$  to the GT segment (introduced in Sec.~\ref{sec:SD}) $\mathbf{\hat{v}} = (\hat{v}_{x}, \hat{v}_{y}, \hat{v}_{w}, \hat{v}_{h}, \hat{v}_{\theta})$:
\begin{equation}
\small
L_{reg} = SmoothedL1(\mathbf{\hat{v}}, \mathbf{v}).
\end{equation} 
Besides, Cross Entropy loss is adopted as $L_{score}$ to train the classifier of SPN in segment detection stage. 

Given $N$ segments, we generate the position-wise combination label map $\mathbf{U}$ of $N \times N$ size for ``combination'' loss. The combination score for position $(p,q)$  can be defined as
\begin{equation}
\small
U_{p,q} = 
\begin{cases}
\textbf{1}, \text{if } P = Q \text{ and } S^{(Q)}_q \in \textit{kNNR}(S^{(P)}_p)  \\
\textbf{0}, \text{otherwise}, \\ 
\end{cases}
\end{equation}
where $S^{(P)}$ and $S^{(Q)}$ represent segments belong to $P$-th and $Q$-th GT polygons. Specifically, if $area(Segment \cap Polygon)/area(Segment) > 0.8$, the segment is deemed belong to the GT polygon. $\textit{kNNR}(S^{(P)}_p)$ denotes a set of adjacent segment pairs to the segment $S^{(P)}_p$ after kNNR applied.
Taking $\mathbf{U}$ as supervision, the position-wise softmax loss is adopted to compute ``combination'' loss $L_{comb}$ between $\mathbf{U}$ and the output feature of last layer of MSGCN. Note, the combination loss at position $(p, q)$ would be ignored if $p$ = $q$ because the relationship between segment and itself is deterministic. 
\section{Experiments}
\subsection{Datasets}

\noindent\textbf{SynthText }~\cite{gupta2016synthetic} is a large scale dataset containing round 800,000 synthetic images. This dataset is utilized to pre-train our model.

\noindent\textbf{MSRA-TD500}~\cite{yao2012detecting} is a dataset collected for detecting long text lines of arbitrary orientations with text line level annotations. It contains 300 images for training and the remaining 200 images constitute the test set.

\noindent\textbf{ICDAR2015} is a dataset proposed as the Challenge 4 of the 2015 Robust Reading Competition~\cite{karatzas2015icdar}  for incidental scene text detection. It consists of 1,000 training images and 500 test images with word level annotations of the quadrangle.

\noindent\textbf{SCUT-CTW1500}~\cite{yuliang2017detecting} is  a challenging dataset collect for detecting curved text. It consists of 1,000 images for training and 500 images for testing, which are labelled by a polygon with 14 points.


\subsection{Implementation Details}

Our method is implemented in Pytorch~\cite{paszke2017automatic}. The network is pre-trained on SynthText for one epoch and fine-tuned on other datasets. Adam~\cite{kingma2014adam} optimizer is adopted for our method, and the learning rate is set to $1e-4$. We use the OHEM proposed in \cite{shrivastava2016training} in training stage to balance training samples and set the ratio of positives to negatives to 1:2. Considering computation cost would be huge if MSGCN takes all segments as input nodes, we apply the Skew-NMS\cite{ma2018arbitrary} after segment detection in both training and inference stage. The threshold of Skew-IoU and angle difference are set to 0.7 and  $\pi/36$ respectively. For data augmentation, we randomly rotate the input images in a certain angle ranging from $-\pi/12$ to $\pi/12$.  Other augmentation tricks, \textit{e.g.,} randomly modifying hue , brightness and contrast, are also adopted. All the experiments are conducted on a regular platform with 8 Nvidia P40 GPUs and 64GB memory. We train our network in batch size of 2 and set it to 1 when evaluating.

\subsection{Ablation Study}\label{sec:abs}

\begin{table}
	\centering
	\small
	\caption{Results of ablation study on SCUT-CTW1500. ``w/o App'', ``w/o Geom'' and ``w/o Graph'' are short for ``without appearance'',  ``without geometry'' and``without graph", respectively.}
	
	\label{tab:ABS}%
	\begin{tabular}{l|c|c|c}
		\hline
		\textbf{Method} & \textbf{Precision} & \textbf{Recall} & \textbf{F-measure}  \\
		\hline
		$\text{PuzzleNet w/o App}$ & 73.4 & 78.0  &75.7\\
		$\text{PuzzleNet w/o Geom}$  & 69.2 & 72.5  &70.8\\
		$\text{PuzzleNet w/o Graph}$  & 65.7 & 69.0  & 67.3\\
		$\text{PuzzleNet-SSim}$  & 80.3  & 83.8 &{82.0}\\
		$\text{PuzzleNet-Square}$ & 73.4 & 78.0  &75.7\\
		$\text{PuzzleNet-MS}$ & 83.3 & \textbf{86.5}  &\textbf{84.9}\\
		\hline
		\textbf{PuzzleNet} &\textbf {84.1}  & \text{84.7} & \text{84.4}\\
		\hline
	\end{tabular}%
	\vspace{-3.5mm}
\end{table}%
In this subsection, we perform several analytic experiments on SCUT-CTW1500 benchmark to verify the effectiveness of our proposed PuzzleNet architecture.  We analyze and investigate the effect of several factors upon the performance, which mainly include context graph mechanism,  segment type and  multi-scale input.  In total we have six variants by training the model based on different combinations of the above factors. In all the experiments using single-scale image as input,  the long side of input image is set to 800. And fine-tuning on  stops at about 100 epochs. The details and corresponding results are shown in Tab.~\ref{tab:ABS}.

\textbf{Effect of Context Graph}\quad In the method ``PuzzleNet w/o App'', we remove the ``appearance'' branch. Now, the module can only capture and exploit the geometry information of segment context to predict their combinations. We observe the F-measure score drops 8.7\%. Comparatively, if we remove the ``geometry'' branch in ``PuzzleNet w/o Geom'' method, we find this setting would be more detrimental to the performance.  In method named ``PuzzleNet w/o Graph'', we completely remove the MSGCN and use three $d$-dimension FC layers instead. For combination prediction, we concatenate appearance and geometry features and compute their Euclidean distance. Under this condition, the graph-based correlation reasoning mechanism is completely removed, we find that simply using appearance and geometry features can not achieve satisfactory results. By comparing the results between ``PuzzleNet'' and ``PuzzleNet-SSim'' which only adopts single cosine similarity function to build graphs, we observe that the ``PuzzleNet'' achieves higher performance due to its ability of ensemble multiple source information from different similarity spaces. From the result ``PuzzleNet'' overtaking above four degraded PuzzleNets by a large margin for all three evaluation protocols, we come to the conclusion that the segment context graph learning mechanism is beneficial to scene text detection performance.

\begin{figure}[htb]
	\begin{center}
		\begin{tabular}{ccc}
			{\hspace{-3pt}}
			
			\includegraphics[width=0.88\linewidth, height = 0.75\linewidth ]{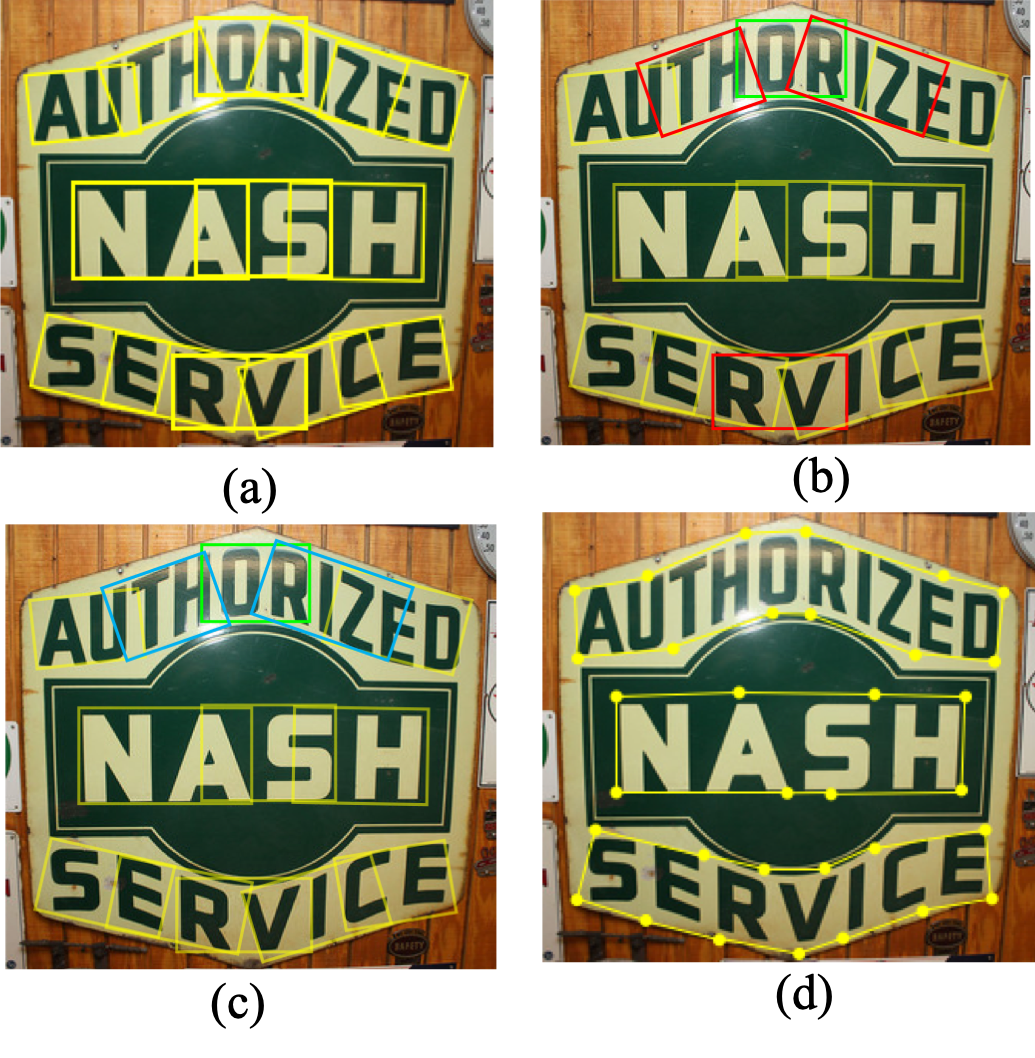}
			\vspace{-4mm}
			
		\end{tabular}
	\end{center}
	\caption{(a) showing the segments predicted by SPN. (b) and
		(c) respectively showing the contextual samples (red) with largest
		appearance similarities and the contextual samples (blue) with largest geometry similarity to the anchor segment (green). Transparent purple segments are the other ones. (d) showing the generated contour points (yellow points) and the linked polygons. } 
	\label{fig:ctxt}
	\vspace{-1mm}
\end{figure}

In Fig.~\ref{fig:ctxt}, for an image from SCUT-CTW1500, we visualize the results output by appearance and geometry similarity modules formulated by Eqn.~\eqref{App} and Eqn.~\eqref{Geom}, which indicate the appearance and geometry contributions of context in combination prediction. For a certain anchor segment (green), its three contextual samples with largest appearance similarities which larger than 0.5 are highlighted in red in Fig.~\ref{fig:ctxt}(b). We observe that they have similar appearance, such as font or background. As to the contextual samples (blue ones in Fig.~\ref{fig:ctxt}(c))with largest geometry similarities which larger than 0.5, we find they locate near the anchor segment, which is reasonable. The above phenomena further demonstrate that our method can extract and aggregate effective context information by a graph model for  text detection task. In addition, the generated contour points introduced in Sec.~\ref{spp} and linked polygons are also visualized in Fig.~\ref{fig:ctxt}(d).

\begin{table}
	\centering
	\small	
	\caption{Results on ICDAR2015.}
	\label{tab:IC15}%
	\begin{tabularx}{\linewidth}{p{2.5cm}|X<{\centering}|X<{\centering}|X<{\centering}}
		\hline
		\textbf{Method} & \textbf{Precision} & \textbf{Recall} & $\textbf{F-measure}$ \\
		\hline
		SegLink~\cite{shi2017detecting}  &73.1 &76.8 &75.0\\
		PixelLink~\cite{deng2018pixellink} &85.5
		&82.0 &83.7\\
		Lyu~\textit{et al.}\cite{lyu2018multi} &{89.5}
		&79.7	&84.3\\
		EAST~\cite{zhou2017east}  &83.3
		&78.3 &80.7\\
		SSTD~\cite{he2017single} & 80.0
		&73.0 &77.0\\
		He \textit{et al.}~\cite{he2017deep} &82.0
		&80.0 &81.0
		\\	
		TextSnake~\cite{long2018textsnake} & 84.9
		&80.4
		&82.6\\
		SPCNET~\cite{xie2019scene} &88.7 &85.8 &87.2\\
		Wang \textit{et al.}~\cite{wang2019arbitrary} &89.2 &86.0 &{87.6}\\
		CRAFT~\cite{baek2019character} &89.8	&84.3	&86.9\\
		PSENet~\cite{wang2019shape} &86.9 &84.5 &{85.7}\\
		FOTS~\cite{liu2018fots} &88.8 &82.0 &85.3 \\ 
		CharNet~\cite{Xing_2019_ICCV} &90.2&81.4&85.6\\
		TextDragon~\cite{Feng_2019_ICCV} &84.8 &81.8 &83.1\\
		LOMO~\cite{zhang2019look} &\textbf{91.3} &83.5 &87.2 \\
		LOMO MS~\cite{zhang2019look} &87.8 &87.6 &\text{87.7} \\
		
		\hline
		\textbf{PuzzleNet} &89.1	&{86.9}	&{88.0}\\
		\textbf{PuzzleNet-MS} &88.9	&\textbf{88.1}	&\textbf{88.5}\\
		\hline
	\end{tabularx}%
\end{table}%

\textbf{Effect of Segment Type}\quad In the method ``PuzzleNet-Square'', we replace the segment representation with the square box used in SegLink~\cite{shi2017detecting}. The comparison results between it and ``PuzzleNet'' verify the local orientation aware rectangular segments can provide more effective information for constructing context graph and achieving higher performance.      

\textbf{Effect of Multi-scale input}\quad By using multi-scale images as input, our PuzzleNet can achieve better recall rate and F-measure, as shown by ``PuzzleNet-MS''. In the multi-scale setting, the long side of  images is resize to $\{400, 600, 800,1000\}$. It demonstrates that multi-scale input could produce more effective segments and our method is capable to combine them in the correct way.


\subsection{Comparison with State-of-the-Art Methods}

\textbf{Results on ICDAR2015}\quad We fine-tune our network on ICDAR2015 in 400 epochs. In single-scale testing, we resize the longer side of input image to 1536. And the longer sides in multi-scale testing are set to $\{1024, 1536, 2048\}$. The comparison of  our method's performance with different state-of-the-arts on ICDAR 2015 are shown in Tab.~\ref{tab:IC15}. With only single-scale testing, our method outperforms most previous methods by a large margin. Specifically, our PuzzleNet can achieve the F-measure of 88.0\%, which surpasses the LOMO MS~\cite{zhang2019look} with multi-scale input. This demonstrates that our proposed PuzzleNet is capable to effectively detect multi-oriented text in complex scenarios.

\begin{table}
	\centering
	\small	
	\caption{Results on MSRA-TD500.}
	\label{tab:TD500}%
	\begin{tabularx}{\linewidth}{p{2.5cm}|X<{\centering}|X<{\centering}|X<{\centering}}
		\hline
		\textbf{Method} & \textbf{Precision} & \textbf{Recall} & $\textbf{F-measure}$  \\
		\hline
		SegLink~\cite{shi2017detecting}  &86.0
		&70.0
		&77.0\\
		Lyu \textit{et al.}~\cite{lyu2018multi} &87.6
		&76.2
		&81.5\\
		EAST~\cite{zhou2017east}  &87.3
		&67.4 &76.1\\
		He \textit{et al.}~\cite{he2017deep} &77.0
		&70.0
		&74.0
		\\	
		TextSnake~\cite{long2018textsnake} & 83.2
		&73.9
		&78.3\\
		PixelLink~\cite{deng2018pixellink}  &83.0 &73.2 &77.8\\
		Wang \textit{et al.}~\cite{wang2019arbitrary}  &85.2 &{82.1} &83.6\\
		CRAFT~\cite{baek2019character} &88.2 &78.2 &82.9\\
		\hline
		\textbf{PuzzleNet} &\textbf{88.2}	&{83.5}	&{85.8}\\
		\textbf{PuzzleNet-MS} &{86.0}	&\textbf{86.2}	&\textbf{86.1}\\
		\hline
	\end{tabularx}%
\end{table}%


\textbf{Results on MSRA-TD500}\quad To further evaluate the performance of our method for detecting long oriented text lines with multi languages, we conduct experiments on MSRA-TD500. After pre-trained on SynthText, our model is fine-tuned in 300 epochs.  In testing, all images are resized to $1280 \times 768$. The quantitative results are shown in Tab.~\ref{tab:TD500}. Our method surpasses all the other methods by a large margin and achieves 2.2\%  gain for F-measure protocol. The results suggest that our method can be readily applied to long text lines with arbitrary orientations in natural images.

\textbf{Results on SCUT-CTW1500}\quad As shown in Tab.~\ref{tab:CTW}, ``PuzzleNet'' using single-scale input image can achieve   84.4\% of the best F-measure compared to other methods. Although LOMO~\cite{zhang2019look} can achieve the best precision, our method overtakes it by 15.1\% for recall protocol. ``PuzzleNet-MS'' adopting multi-scale image can further boost the recall to 86.5\%. The superior performance verifies our method can handle well curved scene text.

\begin{table}
	\centering
	\small	
	\caption{Results on SCUT-CTW1500.}
	\label{tab:CTW}%
	\begin{tabularx}{\linewidth}{p{2.5cm}|X<{\centering}|X<{\centering}|X<{\centering}}
		\hline
		\textbf{Method} & \textbf{Precision} & \textbf{Recall} & $\textbf{F-measure}$ \\
		\hline
		SegLink~\cite{shi2017detecting} & 42.3 &40.0 &40.8\\
		EAST~\cite{zhou2017east}  &78.7 &49.1 &60.4\\
		TextSnake~\cite{long2018textsnake} &67.9 &{85.3} &75.6\\ 
		CTD~\cite{yuliang2017detecting} &74.3 &65.2 &69.5\\
		CTD+TLOC~\cite{yuliang2017detecting} &77.4 &69.8 &73.4\\
		CTPN~\cite{tian2016detecting}  & 60.4 & 53.8  &56.9\\
		DMPNet~\cite{liu2017deep} &69.9 &56.0 &62.2\\
		Wang \textit{et al.}~\cite{wang2019arbitrary} &80.1 &80.1 &80.2\\
		CRAFT~\cite{baek2019character} &86.0 &81.1  &83.5\\
		PSENet~\cite{wang2019shape} &84.8 &79.7 &{82.2}\\
		TextDragon~\cite{Feng_2019_ICCV} & 84.5 &82.8 &83.6\\
			
		LOMO~\cite{zhang2019look} &\textbf{89.2} &69.6 &78.4 \\
		LOMO MS~\cite{zhang2019look} &85.7 &76.5 &80.8 \\
		\hline
		\textbf{PuzzleNet} & {84.1}  & \text{84.7} & \text{84.4}\\
		\textbf{PuzzleNet-MS} & 83.3 & \textbf{86.5}  &\textbf{84.9}\\
		\hline
	\end{tabularx}%
\end{table}%

%
%

\subsection{Detected Examples on Benchmark Datasets}
\begin{figure}[htb]
	\begin{center}
		\begin{tabular}{ccc}
			{\hspace{-3pt}}
			
			\includegraphics[width=0.85\linewidth]{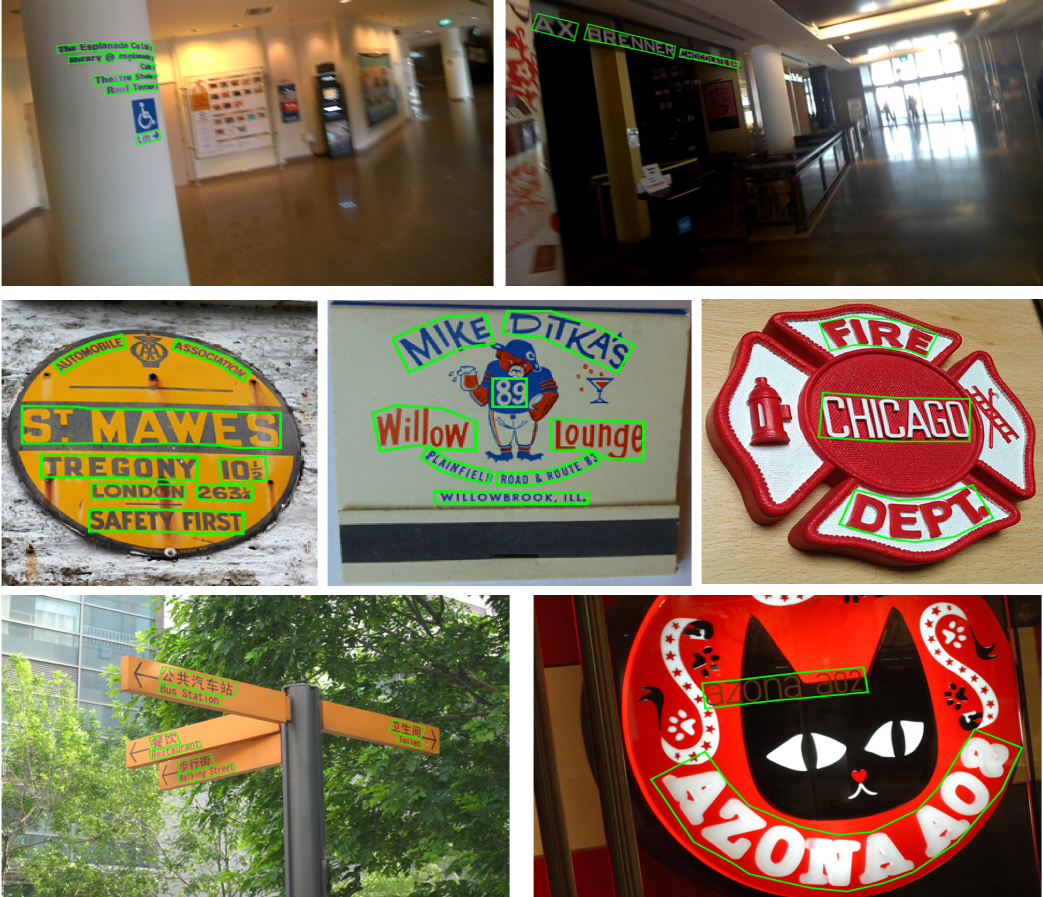}
			\vspace{-1mm}
			
		\end{tabular}
	\end{center}
	\caption{Examples of detection results. From top to bottom in rows: ICDAR2015, SCUT-CTW1500, MSRA-TD500 } 
	\label{fig:examp}
	\vspace{0mm}
\end{figure}

In Fig.~\ref{fig:examp}, we also visualize detection results
produced by our PuzzleNet for testing samples from ICDAR2015, SCUT-CTW1500 and MSRA-TD500 dataset. The first row is from ICDAR2015, while the second and third rows are from SCUT-CTW1500 and MSRA-TD500 respectively. We observe that our detection model can predict precise description of the shape and course of text instances with arbitrary orientations and shapes. We attribute such ability to the segment context graph learning mechanism.

\section{Conclusions}
We present a novel decomposition-based method solves scene text detection through learning segment context graph. Extensive experiments
on three public benchmarks demonstrate its superiority over state-of-the-art methods in most cases in scene text detection.

{\small
\bibliographystyle{ieee_fullname}
\bibliography{detbib}
}

\end{document}